\documentclass[letterpaper, 10 pt, conference]{ieeeconf}  % Comment this line out if you need a4paper

\IEEEoverridecommandlockouts           % This command is only needed if 
\usepackage{amsmath,amsfonts}
\usepackage{algorithmic}
\usepackage{algorithm}
\usepackage{array}
\usepackage[caption=false,font=normalsize,labelfont=sf,textfont=sf]{subfig}
\usepackage{textcomp}
\usepackage{stfloats}
\usepackage{url}
\usepackage{verbatim}
\usepackage{graphicx}
\usepackage{cite}
\usepackage{multirow} 
\usepackage{pifont}
\newcommand{\cmark}{\ding{51}} % 定义勾号符号
\usepackage{array}
\newcommand{\Tstrut}{\rule{0pt}{2.6ex}} % 表格顶部空白
\newcommand{\Bstrut}{\rule[-0.9ex]{0pt}{0pt}} % 表格底部空白
\usepackage{xcolor}
\usepackage{colortbl}
\usepackage{caption}
\usepackage{hyperref}
\title{\LARGE \bf
Robust Unsupervised Domain Adaptation for\\ 3D Point Cloud Segmentation Under Source Adversarial Attacks
}

\author{
Haosheng Li\textsuperscript{1}\textsuperscript{\textsection},
Junjie Chen\textsuperscript{1}\textsuperscript{\textsection},
Yuecong Xu\textsuperscript{2},
Kemi Ding\textsuperscript{1}
\thanks{H. Li, J. Chen and K. Ding are with the Department of Automation and Intelligent Manufacturing (AIM), Southern University of Science and Technology, Shenzhen, China. Email: {\tt\small\{12332662, 12332651\} @mail.sustech.edu.cn, dingkm@sustech.edu.cn.}}
\thanks{Y. Xu is with the Department of Electrical and Computer Engineering, National University of Singapore. Email: 
{\tt\small yc.xu@nus.edu.sg}}}

\begin{document}

\hbadness=2000000000
\vbadness=2000000000
\hfuzz=100pt

\setlength{\abovedisplayskip}{3pt}
\setlength{\belowdisplayskip}{3pt}
\setlength{\floatsep}{3pt plus 1.0pt minus 1.0pt}
\setlength{\intextsep}{3pt plus 1.0pt minus 1.0pt}
\setlength{\textfloatsep}{3pt plus 1.0pt minus 1.0pt}
\setlength{\parskip}{0pt}

\maketitle
\begingroup\renewcommand\thefootnote{\textsection}
\footnotetext{Equal contribution.}
\endgroup
% \thispagestyle{empty}
% \pagestyle{empty}

%%%%%%%%%%%%%%%%%%%%%%%%%%%%%%%%%%%%%%%%%%%%%%%%%%%%%%%%%%%%%%%%%%%%%%%%%%%%%%%%
\begin{abstract}
Unsupervised domain adaptation (UDA) frameworks have shown good generalization capabilities for 3D point cloud semantic segmentation models on clean data. However, existing works overlook adversarial robustness when the source domain itself is compromised. To comprehensively explore the robustness of the UDA frameworks, we first design a stealthy adversarial point cloud generation attack that can significantly contaminate datasets with only minor perturbations to the point cloud surface. Based on that, we propose a novel dataset, AdvSynLiDAR, comprising synthesized contaminated LiDAR point clouds. With the generated corrupted data, we further develop the Adversarial Adaptation Framework (AAF) as the countermeasure. Specifically, by extending the key point sensitive (KPS) loss towards the Robust Long-Tail loss (RLT loss) and utilizing a decoder branch, our approach enables the model to focus on long-tail classes during the pre-training phase and leverages high-confidence decoded point cloud information to restore point cloud structures during the adaptation phase. We evaluated our AAF method on the AdvSynLiDAR dataset, where the results demonstrate that our AAF method can mitigate performance degradation under source adversarial perturbations for UDA in the 3D point cloud segmentation application.

\end{abstract}

%%%%%%%%%%%%%%%%%%%%%%%%%%%%%%%%%%%%%%%%%%%%%%%%%%%%%%%%%%%%%%%%%%%%%%%%%%%%%%%%
\section{Introduction}

{I}{n} applications such as autonomous driving and robotic navigation~\cite{10403932}, 3D point cloud semantic segmentation (PCSS)~\cite{9309360} is a critical technology. Due to the complexity of 3D data and the high annotation cost, unsupervised domain adaptation (UDA) for PCSS has emerged as an effective strategy for transferring knowledge from labeled source domain to an unlabeled target domain. While existing UDA methods perform well on clean data, their robustness against adversarial source perturbations remains unexplored. In real-world scenarios, source domain data used for transfer can be compromised by subtle adversarial manipulations, such as geometry distortion or label noise. Although these perturbations may appear as minor noise visually, they can significantly degrade model performance, inducing erroneous decisions in robot systems. 

In the context of cross-domain adaptation, current research lacks datasets subjected to adversarial attacks, thus leading to the absence of effective defense solutions for cross-domain 3D point cloud segmentation. To comprehensively evaluate the robustness of the UDA framework under adversarial conditions, we introduce the AdvSynLiDAR dataset based on SynLiDAR \cite{xiao2022transfer}, simulating scenarios where the source domain data undergoes severe adversarial attacks. We introduce a dynamic adjustment factor into the PGD attack \cite{madry2017towards}, modulating the perturbation intensity according to the distance from the point to the viewpoint center. Further,
% to simulate the challenges encountered during data collection and labeling in real-world scenarios, 
we incorporate high-confidence incorrect labels into the training process, leading to more invisible errors. In the face of such attacks, our task faces two significant difficulties: first, the stealthiness of adversarial attacks makes them difficult to be detected; second, data attacks needs to be defended while considering their long-tail properties simultaneously. These difficulties present both a challenge and an opportunity. We observe that for long-tail data which is common in 3D point clouds, despite its scarcity, it also has a much lower probability of being targeted by attacks. This offers a potential advantage to leverage upon long-tail data for robust cross-domain segmentation.

\begin{figure}[t]
    \centering
    \includegraphics[width=0.95\linewidth]{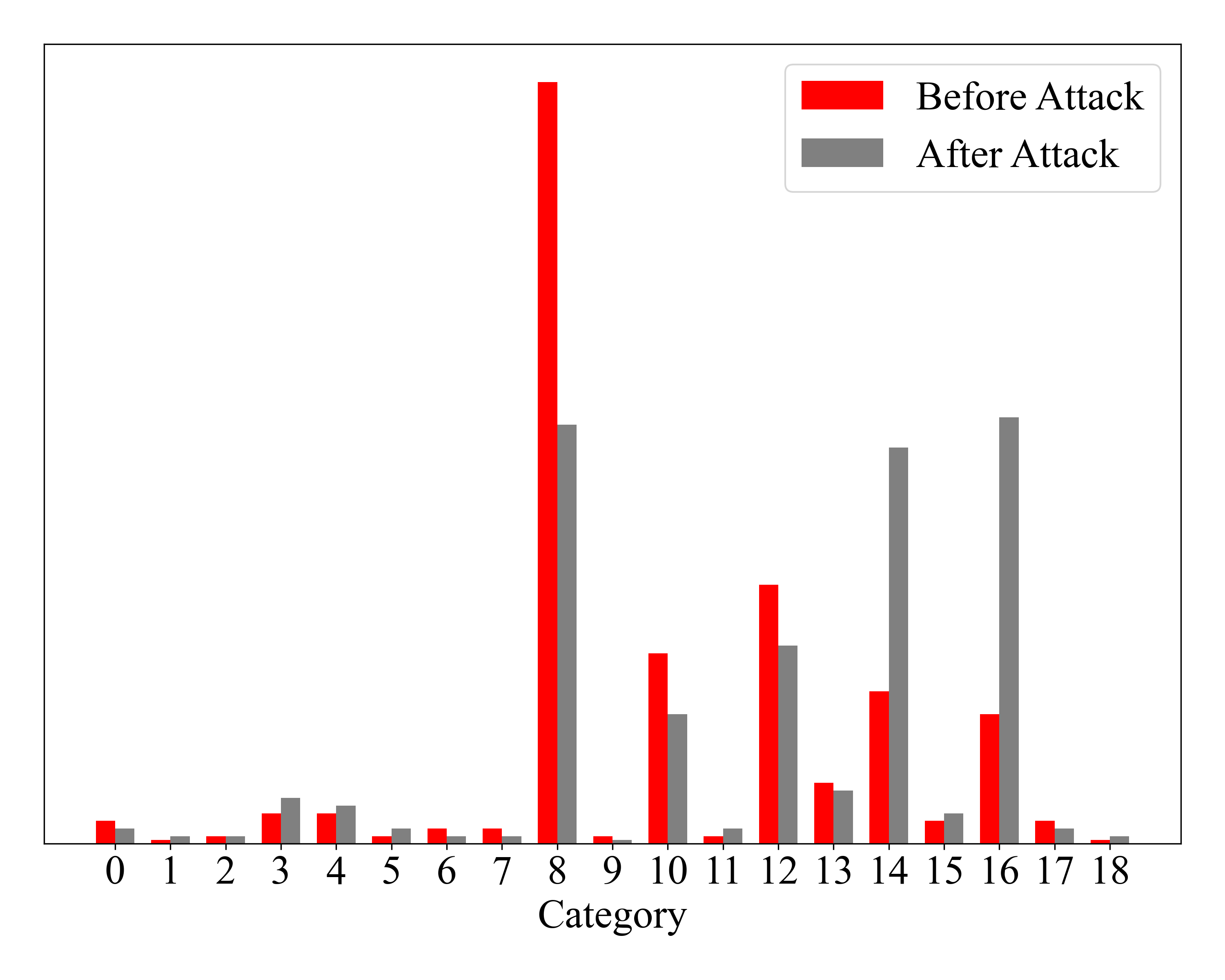}
    \caption{Illustration of the distribution of various classes within the 3D point cloud data before and after adversarial attacks of SemanticKITTI. The long-tail classes exhibit relatively stable distribution, with minimal changes observed across the pre-attack and post-attack states.}
    % \vspace{-5mm}
    \label{Fig: Intro}
\end{figure}

To this end, we propose the Adversarial Adaptation Framework (AAF) defense strategy. Our approach is divided into the pre-training and adaptation phases. During the pre-training phase, we focus mainly on handling long-tail classes. As shown in Fig.~\ref{Fig: Intro}, long-tail classes exhibit stronger robustness against attacks than head classes. Inspired by the KPS loss \cite{li2022key} in 2D classification tasks, we introduce the Robust Long-Tail Loss (RLT loss) to enhance the model's attention to 3D long-tail class data. We first redefine the key points in KPS loss based on the characteristics of 3D point clouds. We then design a new margin adjustment mechanism that dynamically adjusts the margins based on the actual distribution of each class to address the spareness and uneven distribution of long-tail classes.

Further, though some existing methods (e.g., sampling or removing surface outliers) can effectively reduce the impact of troubled point clouds on model predictions, these methods cannot fully restore surface deformations or spatial distribution of point clouds when facing stealthy disturbances. Inspired by the VRCNet~\cite{pan2021variational} point cloud completion network, we introduce a decoder branch to restore point clouds' distribution and spatial information during the adaptation phase. This approach uses probabilistic models combined with the decoded feature distribution of the input point clouds to restore and generate fine-grained point clouds consistent with the distribution of normal point cloud data. Moreover, by combining the restored point cloud with high-confidence neighborhood semantic information from the original point cloud, we generate pseudo-labels that enhance the model's prediction stability, which is used for model training during the transfer phase, thus improving the semantic segmentation performance in the target domain after adaptation without detecting the stealthiness of attacked data.

In summary, our contributions are threefold. (i) To the best of our knowledge, our research is the first to explore cross-domain robustness and defense strategies in 3D point cloud semantic segmentation. (ii) Our defense strategy AAF improves the model's robustness by introducing the RLT loss during the pre-training phase to enhance the focus on long-tail classes and effectively utilizing high-confidence decoded point cloud information during the adaptation phase. (iii) Experimental results demonstrate that AAF achieves a remarkable improvement on the SemanticKITTI and SemanticPOSS datasets after the attack, with an increase of 11.61 and 9.85 mIoU, respectively.
\section{Related Work}

\subsection{Unsupervised Domain Adaptation}
It is often observed that the data distribution between the source and target domains diverge, resulting in a notable decline in performance when directly applying models trained on the source domain to the target domain.
To address the domain shift between source and target datasets in the unsupervised settings, domain adaptation methods aim to bridge the distribution gap at various levels. Input-level adaptation \cite{9804815,Hoffman_cycada2017} aligns the visual appearance of source and target images using style transfer, ensuring domain uniformity and semantic consistency for training. Feature-level adaptation \cite{10737391,8578493} aligns the latent representations of source and target domains to extract domain-invariant features, enabling the network to segment both domains from a shared feature space. Output-level adaptation \cite{spadotto2021unsupervised,8954439} aligns the segmentation prediction maps across domains, often using adversarial strategies and target label statistics to enhance cross-domain performance.

\subsection{Point Cloud Adversarial Attacks and Countermeasures}
Various methods have been proposed to generate adversarial attacks on 3D point cloud models. Gradient-based strategies \cite{kim2021minimal} utilize the gradient information of the model and iteratively adjust the input to maximize the loss function for creating adversarial perturbations. Optimization strategies \cite{zhang20233d} focus on the model output logits to develop attacks. Other structure-based attacks include point shift attacks \cite{tang2022normalattack}, point add attacks \cite{shi2022shape}, point drop attacks \cite{naderi2022model}, etc. While the preceding attacks have proven effective for classification and segmentation tasks, their impact on cross-domain adaptation, particularly in transferring semantic information across samples under adversarial perturbations in the source domain, remains largely unexplored.

Countermeasures against 3D point cloud adversarial attacks mainly fall into data-driven \cite{naderi2023lpf} and model-driven \cite{liu2022imperceptible} approaches. For data-driven approaches, PointGuard \cite{liu2021pointguard} classifies random subsets of a point cloud and uses majority voting to make the final prediction. Meanwhile, DUP-Net \cite{zhou2019dup} combines statistical outliers removal with an up-sampling network, aiming to denoise before up-sampling to restore the original shape of the point cloud. For model-driven approaches, LPC~\cite{li2022robust} leverages centroid coordinates on a lattice to transform point clouds into two-dimensional images, enhancing the model's resilience to adversarial attacks through structured sparse encoding and a two-layer optimization framework. However, to the best of our knowledge, existing research has not addressed defense mechanisms for cross-domain segmentation tasks.

\section{The AdvSynLiDAR Dataset}
% \subsection{Background}
Generally, point clouds in the source domain are susceptible to subtle perturbations and deformations, which can be caused by sensor noise, environmental variations, and preprocessing artifacts~\cite{xu2023adversarial,Gao_2023_ICCV}.
This can significantly degrade model performance and increase distribution discrepancies across domains, leading to inferior 3D unsupervised cross-domain adaptation (UDA) performance for point cloud semantic segmentation. Despite the significance of this challenge, to the best of our knowledge, there is currently a lack of standardized methods for evaluating the impact of such perturbations during adaptation. To address these issues, we introduce the AdvSynLiDAR dataset, built upon the structured SynLiDAR dataset~\cite{xiao2022transfer}.
Specifically, our dataset focuses on understanding the impact of attacks on the source domain. 

\subsection{Imperceptible Perturbation Points}
In real-world scenarios, adversarial perturbations are usually subtle and spatially dependent, making them difficult to be detected. To mimic realistic attacks, we propose a perceptually-aware perturbation strategy. This strategy applies more significant perturbations to distant regions while minimizing changes in dense, perceptually sensitive areas, enabling a more precise evaluation of UDA robustness under an adversarial source domain.

\begin{figure}[t]
    \centering
    \includegraphics[width=1.\linewidth]{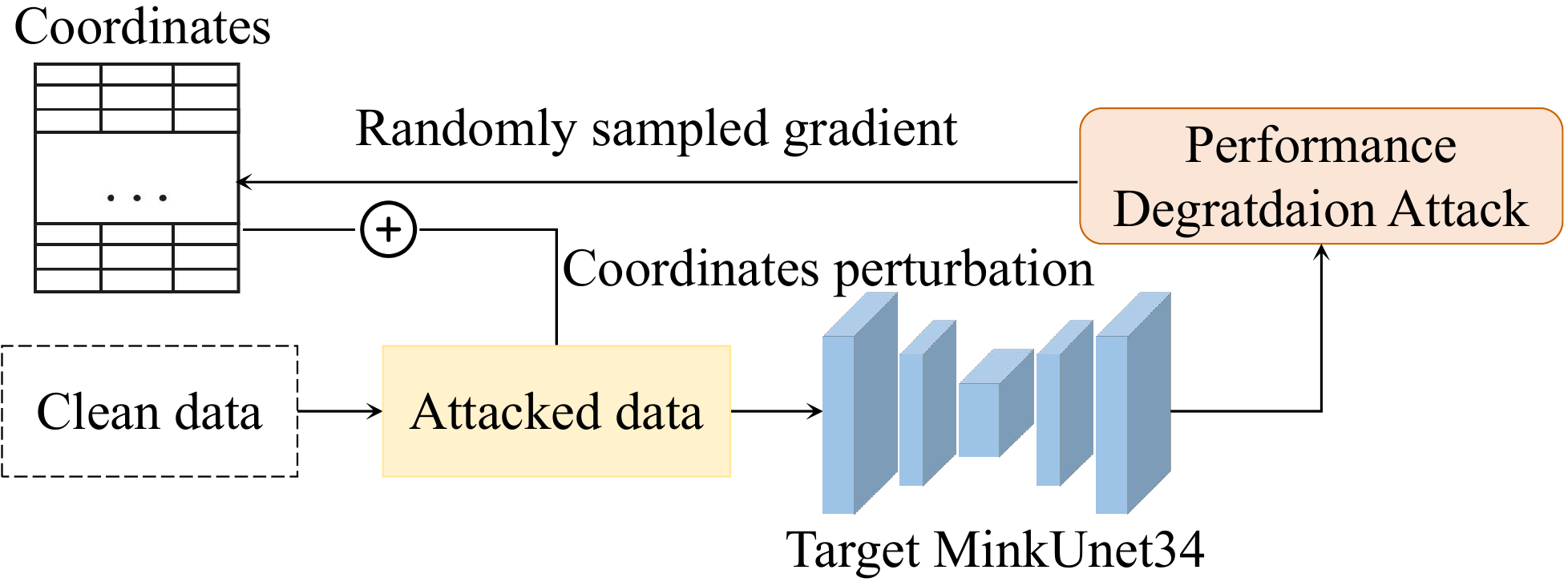}
    \caption{Generating adversarial source domain coordinates, $\gamma^{(i)}$ controls the intensity of perturbations based on the distance to the viewpoint center.}
    \label{Fig:adversarial_sample_generation}
    % \vspace{-3mm}
\end{figure}

\subsection{Noisy Semantic Labels}
Manual annotations in 3D point clouds often contain label noise, especially for occluded, distant, or complex objects, where annotation errors occur more frequently~\cite{ye2021learning}. 
While standard UDA approaches assume that source domain annotations are reliable, real-world scenarios often involve systematic annotation noise, where annotators or automated labeling systems confidently assign incorrect labels. To systematically evaluate the impact of such errors, we introduce high-confidence mislabeling into the source domain data. This approach generates the highest-confidence incorrect class as the adversarial target, mimicking real-world scenarios with noisy labels. 

\section{Methodology}
To tackle the vulnerability of cross-domain point cloud segmentation under adversarial perturbed source domain, we propose an Adversarial Adaptation Framework (AAF), which improves robustness by enforcing resilient training objectives and preserving semantic consistency under source-domain perturbations.

\subsection{Preliminary and Overview}
In the context of adversarial cross-domain adaptation, the labeled source domain is denoted as \( \mathcal{S} = \{(\mathbf{x}_{i}, \mathbf{y}_{i})\}_{i=1}^{|\mathcal{S}|} \). Here, \( \mathbf{x}_{\mathcal{S}} \in \mathbb{R}^{N \times 3} \) represents the 3D point cloud coordinates of the source domain. The label \(\mathbf{y}_{\mathcal{S}}^{(j)} \in \mathbf{Y}_\mathcal{S} = \{0, 1, \ldots, c-1\} \) denotes the semantic label for each point, where $\mathbf{Y}_\mathcal{S}$ is the label space of point clouds across domains, and \( c \) is the number of classes. The objective of our cross-domain adaptation task is to extract semantic information from the source domain and transfer it to an unlabeled target domain, which is denoted as $\mathcal{T} = \{(\mathbf{x}_{i})\}_{i=1}^{|\mathcal{T}|}$. To obtain a point-wise classification of point clouds, the pre-trained model $\theta$ provides the probability distribution of the $j$-th point in a source point cloud $\mathbf{x}_{\mathcal{S}}$, which is denoted as $\Phi_{\theta}(\mathbf{x}_{\mathcal{S}})^{(j)}=p(\cdot|\mathbf{x}_{\mathcal{S}})^{(j)} \in \mathbb{R}^{|c|}$.

\subsection{Adversarial Adaptation Framework (AAF)}

\subsubsection{Robust Long-Tail Loss (RLT Loss)}
% As shown in Fig~\ref{Fig: warmup},
During the pre-training phase, in order to balance the advantages of KPS loss \cite{li2022key} in handling long-tail classes with the strengths of SoftDICELoss in managing head classes, we introduce the Robust Long-Tail loss (RLT loss). SoftDiceLoss is widely used in segmentation tasks to measure the overlap between predictions and ground truth labels. Initially designed for 2D classification tasks, KPS loss aims to enhance the separability of features by adjusting the boundaries between key points and non-key points in the feature space. To apply KPS loss to higher-dimensional 3D segmentation tasks, we make several improvements to the original KPS loss:

\noindent \textbf{Geometric Importance in 3D Key Point Definition.} We introduce geometric importance as the criteria for defining key points in 3D point clouds, allowing for better identification of critical points under attack conditions. This enhances the discriminative ability of KPS loss for long-tail classes.

\begin{figure*}[t]
    \centering
    \includegraphics[width=0.8\linewidth]{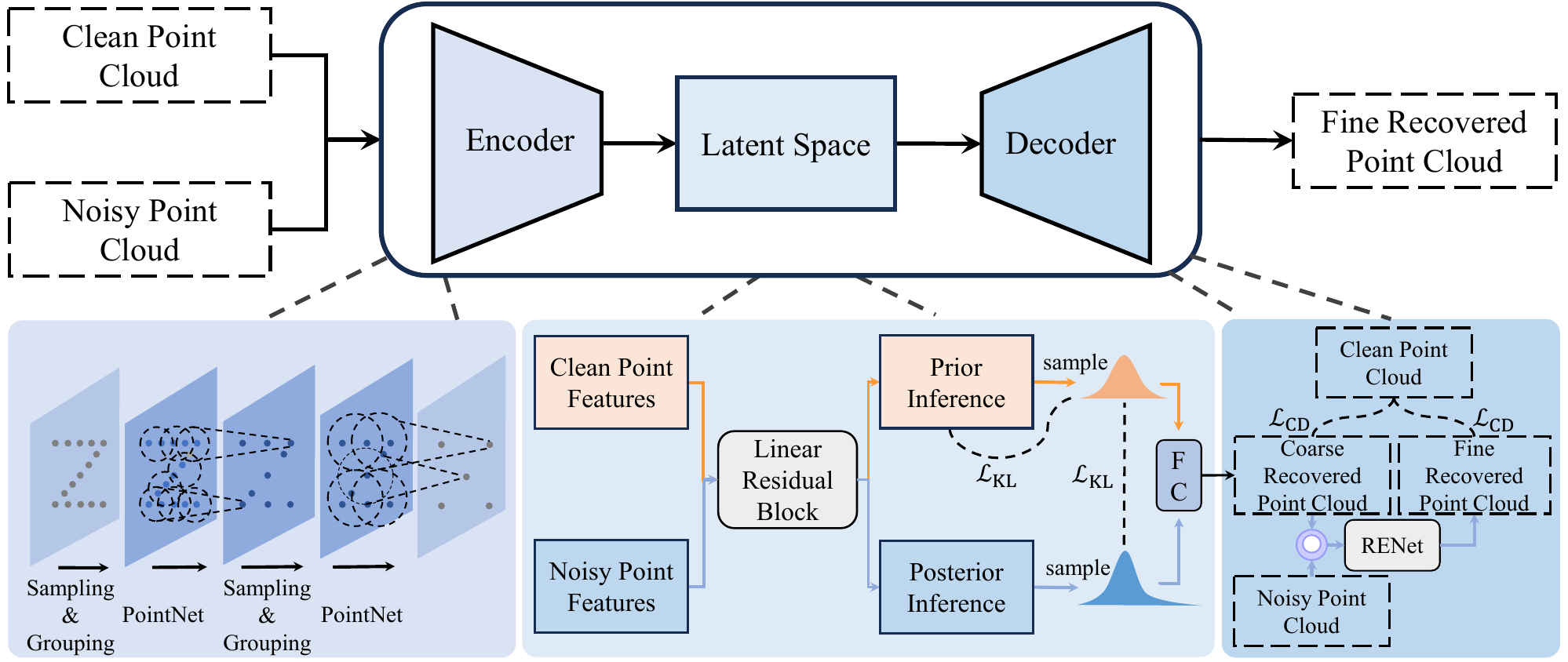}
    \caption{The figure illustrates the pre-training process of the decoder, where the symmetric Chamfer Distance $\mathcal{L}_\mathbf{CD}$ quantifies the discrepancy between the reconstructed and original point clouds. Additionally, the Kullback-Leibler divergence loss $\mathcal{L}_\mathbf{KL}$ aligns the probability distribution from prior and posterior inference, promoting consistency in the latent space.}
    \label{Fig:Decoder}
    % \vspace{-3mm}
\end{figure*}

\noindent \textbf{Dynamic Boundary Adjustment for Class Imbalance.} In standard classification tasks, all classes share the same decision boundary, which may not be ideal for long-tail classes due to their sparse distribution and low sample count. However, 3D point clouds often exhibit severe class imbalance, especially for long-tail classes as depicted in Fig.~\ref{Fig: Intro}. To address this issue, we introduce a dynamic boundary adjustment mechanism that adjusts each class's boundary based on sample density and classification difficulty. In this way, the smaller the class size (i.e., long-tail classes), the larger the boundary adjustment term $m_{\mathbf{y}_i}$, giving these classes more decision space during classification. 

In summary, the KPS loss in 3D tasks encourages more significant boundaries for key points, thereby promoting the separability of long-tail class features.

The overall loss function $\mathcal{L}_{\text{RLT}}$ , can thus be defined as:
\begin{equation}
\mathcal{L}_{\text{RLT}}=\lambda\cdot\mathcal {L}_{\text {KPS }}+(1-\lambda)\cdot\mathcal {L}_{\text {SD }},
\end{equation}
where we employ Bayesian optimization \cite{frazier2018tutorial} to dynamically adjust the regularization parameter $\lambda$ in the loss function, aiming to identify the optimal $\lambda$ at each iteration and prevent overfitting.

\subsubsection{Probability Decoder Branch} 
The probability decoder branch is introduced to work in tandem with the adaptation model, which consists of two key components: a probability decoder that aligns the input distribution with the distribution of clean source domain and a High-confidence Nearest Pseudo Update (HNPU) strategy that aggregates high-confidence information. These approaches help to improve model robustness under adversarial source point clouds. 

\noindent $\textbf{Probability Decoder.}$ Inspired by VRCNet~\cite{pan2021variational}, we introduce a probability decoder to mitigate latent discrepancies in source inputs dynamically. As illustrated in Fig.~\ref{Fig:Decoder}, to simulate realistic noise, we generate perturbed inputs by adding Gaussian noise and random augmentations to the clean source coordinates.

The decoder branch reconstructs the clean point cloud in three stages: reconstruction, completion, and enhancement, which progressively restore global and local geometric details. Specifically, in the reconstruction stage, VRCNet encodes the input features into global features and the latent distribution. It then reconstructs a coarse point cloud using the decoding distribution. 

Additionally, the parameter $\lambda$ serves as a weighting factor. During the completion process, the input point cloud is reconstructed using the global feature and the latent variable distribution derived from the perturbed and augmented input point cloud $\mathbf{x}_g$. During training, the posterior distribution is regularized to approach the prior distribution. 

In the enhancement stage, a self-attention module and up-sampling layers are added to capture local features at different scales and restore the point cloud to finer granularity. This enables adaptive restoration of perturbed point clouds while preserving essential structural information, mitigating the risk of introducing additional noise or erroneous labels.

\begin{figure}[t]
    \centering
    \includegraphics[width=1.\linewidth]{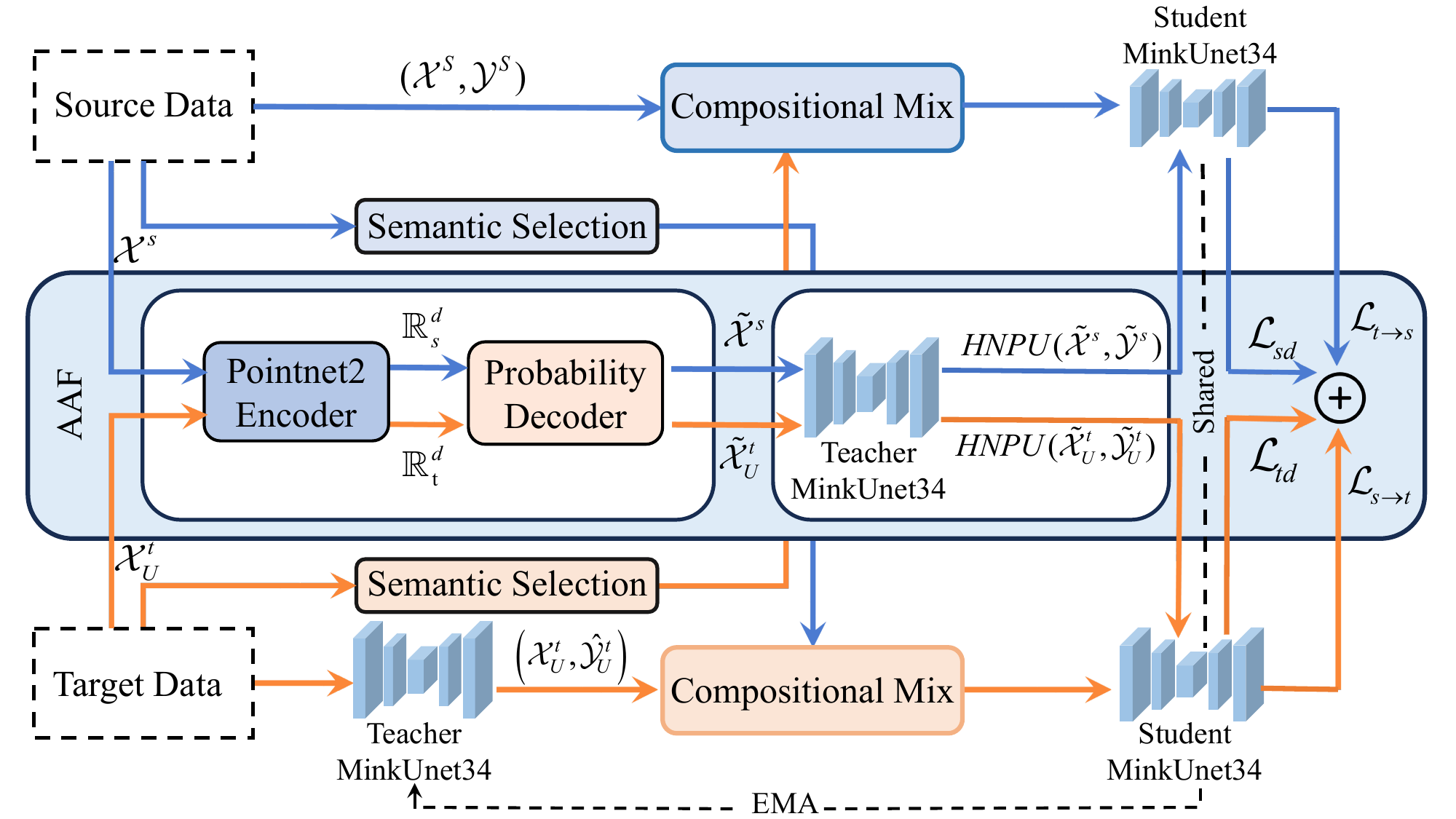}
    \caption{Illustration of cross-domain adaptation with the AAF framework. The probability decoder branch dynamically adjusts input distributions while HNPU enhances pseudo generation throughout the adaptation process.}
    \label{Fig: adversarial cross-domain framework}
\end{figure}

\noindent \textbf{High-Confidence Nearest Pseudo Update.} Motivated by existing work, the robustness of the model can be further strengthened by concentrating on high confidence areas~\cite{pmlr-v80-wu18e}. We introduce the High-confidence Nearest Pseudo Update (HNPU) Algorithm~\ref{alg:hnpu} to aggregate information from high-confidence neighborhoods for more robust adaptation.

As shown in Fig.~\ref{Fig: adversarial cross-domain framework}, a teacher-student network transfers information between the source and target domains during the cross-domain adaptation. This is done using the adaptation loss $\mathcal{L}_{tot} = \mathcal{L}_{s\to t} + \mathcal{L}_{t\to s}$, which captures consistent semantic information between the source and target domains. In the decoder branch, the adaptation process incorporates additional segmentation losses introduced by the decoded coordinates and the pseudo-labels updated by HNPU. 

Intuitively, the decoder branch aligns the input distribution with the clean source distribution. The probability decoder transfers spatial information from the source domain, helping to mitigate adversarial shifts in input point clouds and reduce the domain gap for target domain data. In contrast, HNPU significantly mitigates the impact of anomalous source domain data, further enhancing the performance of cross-domain adaptation.

\section{Experiments}
In this section, we evaluate the performance of our proposed Adversarial Adaptation Framework (AAF) through adversarial cross-domain semantic segmentation experiments, explicitly focusing on the synthetic-to-real domain shift. We present class-wise semantic segmentation results for two distinct scenarios and provide a detailed comparison to illustrate the effectiveness of our approach. Additionally, ablation studies and empirical analysis are also presented to justify the design of AAF.

\subsection{Experimental Settings}
We perform adversarial cross-domain adaptation for 3D semantic segmentation tasks on SynLiDAR~\cite{xiao2022transfer}, SemanticPOSS~\cite{pan2020semanticposs}, and SemanticKITTI~\cite{behley2019semantickitti} datasets. SynLiDAR is a large-scale synthetic dataset created using Unreal Engine comprising 198,396 annotated point clouds with 32 semantic classes. SemanticPOSS includes 2,988 real-world point clouds categorized into 14 semantic classes. SemanticKITTI is a comprehensive segmentation dataset derived from LiDAR acquisitions of the KITTI dataset, consisting of 43,552 annotated real-world point clouds with over 19 semantic classes. Following CosMix~\cite{saltori2023compositional}, we adopt the same training and validation splits. 

\subsection{Performances and Comparisons}
Tables. \ref{table:4-1 lidar2kitti} and \ref{table:4-2 lidar2poss} present class-wise cross-domain semantic segmentation results on SynLiDAR $\rightarrow$ SemanticKITTI and SynLiDAR $\rightarrow$ SemanticPOSS, respectively.

\renewcommand{\arraystretch}{1.3}
\begin{table*}[t]
\center
\caption{Results for class-wise semantic segmentation on SynLiDAR to SemanticKITTI. Selection-perc is the hyperparameter that regulates the ratio of selected classes. Source classes are selected randomly with a selection-perc of 0.5 originally.}
\Large % Reduce font size
\resizebox{0.9\linewidth}{!}{
\begin{tabular}{c|c|c|ccccccccccccccccccc>{\columncolor{gray!30}}c}
\hline
\hline
  \multirow{1}{*}{\textbf{Source}} &
  \multicolumn{1}{c|}{\parbox[t]{2mm}{\rotatebox{90}{sel-p.}}} & 
  \multicolumn{1}{c|}{\parbox[t]{2mm}{\rotatebox{90}{AAF}}} & 
  \multicolumn{1}{c|}{\parbox[t]{2mm}{\rotatebox{90}{car}}} & 
  \multicolumn{1}{c|}{\parbox[t]{2mm}{\rotatebox{90}{bi.cle}}} & 
  \multicolumn{1}{c|}{\parbox[t]{2mm}{\rotatebox{90}{mt.cle}}} & 
  \multicolumn{1}{c|}{\parbox[t]{2mm}{\rotatebox{90}{truck}}} & 
  \multicolumn{1}{c|}{\parbox[t]{2mm}{\rotatebox{90}{oth-v.}}} & 
  \multicolumn{1}{c|}{\parbox[t]{2mm}{\rotatebox{90}{pers.}}} & 
  \multicolumn{1}{c|}{\parbox[t]{2mm}{\rotatebox{90}{b.clst}}} & 
  \multicolumn{1}{c|}{\parbox[t]{2mm}{\rotatebox{90}{m.clst}}} & 
  \multicolumn{1}{c|}{\parbox[t]{2mm}{\rotatebox{90}{road}}} & 
  \multicolumn{1}{c|}{\parbox[t]{2mm}{\rotatebox{90}{park.}}} & 
  \multicolumn{1}{c|}{\parbox[t]{2mm}{\rotatebox{90}{sidew.}}} & 
  \multicolumn{1}{c|}{\parbox[t]{2mm}{\rotatebox{90}{oth-g.}}} & 
  \multicolumn{1}{c|}{\parbox[t]{2mm}{\rotatebox{90}{build.}}} & 
  \multicolumn{1}{c|}{\parbox[t]{2mm}{\rotatebox{90}{fence}}} & 
  \multicolumn{1}{c|}{\parbox[t]{2mm}{\rotatebox{90}{veget.}}} & 
  \multicolumn{1}{c|}{\parbox[t]{2mm}{\rotatebox{90}{trunk}}} & 
  \multicolumn{1}{c|}{\parbox[t]{2mm}{\rotatebox{90}{terra.}}} & 
  \multicolumn{1}{c|}{\parbox[t]{2mm}{\rotatebox{90}{pole}}} & 
  \multicolumn{1}{c|}{\parbox[t]{2mm}{\rotatebox{90}{traff.}}} & 
  \multicolumn{1}{c|}{\parbox[t]{2mm}{\rotatebox{90}{mIoU}}}\Tstrut\Bstrut\\
\hline
\multirow{2}{*}{SynLiDAR} & 0.5 &
& 79.30	& 9.12 & 30.68 & 20.88 & 11.54 & 24.91 & 27.00 &	17.62 &	78.03 &	14.65 &	47.12 & 0.19 & 53.52 & 13.66 & 68.72 & 32.75 & 31.94 & 38.17 &	13.96 & 32.30\\
& 0.9 &
& 80.07 & 5.46 & 25.70 & 30.32 & 10.86 & 23.32 & 21.51 &	7.77 & 80.04 & 18.70 & 49.36 & 0.15 & 49.45 & 13.46 & 67.26 & 29.56 & 34.01 & 40.04 & 12.46 & 31.55\\
& 0.5 & \cmark & 77.51 & 6.59 & 30.51 & 19.01 & 9.45 & 24.83 & 29.26 & 18.93 & 77.72 & 13.79 & 45.94 & 0.09 & 50.09 & 13.74 & 68.05 & 28.27 & 29.20 & 37.76 & 14.00 & 31.30\\
\hline
\multirow{2}{*}{AdvSynLiDAR} 
& 0.5 & & 43.22 & 0.72 & 5.96 &	5.65 & 2.83 & 18.50 & 4.24 & 9.17 &	42.57 &	1.33 & 8.88 & 1.15 & 36.71 & 9.20 & 35.90 & 19.29 & 5.24 & 33.28 & 6.47 &15.28 \\
& 0.9 & & 40.50	& 9.47 & 6.46 &	10.39 & 4.94 & 20.74 & 28.77 & 15.50 & 46.65 & 3.70 & 14.00 & 1.51 & 43.09 & 10.74 & 41.66 & 23.66 & 6.67 & 38.48 & 10.53 & 19.87\\
& 0.5 & \cmark & 55.32 & 5.16 & 4.52 & 15.78 & 6.11 & 14.26 & 19.75 & 7.72 & 56.92 & 6.79 & 16.77 & 1.09 & 41.12 & 10.90 & 61.21 & 25.63 & 7.46 & 36.42 & 8.12 & 22.50\\
\hline
\hline
\end{tabular}
}
% \smallskip
\label{table:4-1 lidar2kitti}
% \vspace{-2mm}
\end{table*}

% SynLiDAR to SemanticKITTI
\renewcommand{\arraystretch}{1.3}
\begin{table*}[t]
\center
% \scriptsize
\caption{Results for class-wise semantic segmentation on SynLiDAR to SemanticPOSS. Source classes are selected randomly with a selection-perc of 0.5 originally.}
\resizebox{0.9\linewidth}{!}{\noindent
\begin{tabular}{c|c|c|ccccccccccccc>{\columncolor{gray!30}}c}
\hline
\hline
  \multirow{1}{*}{\textbf{Source}} &
  \multicolumn{1}{c|}{\centering\parbox[t]{2mm}{\rotatebox{90}{sel-p.}}} & 
  \multicolumn{1}{c|}{\parbox[t]{2mm}{\rotatebox{90}{AAF}}} & 
  \multicolumn{1}{c|}{\parbox[t]{2mm}
  {\rotatebox{90}{pers.}}} & 
  \multicolumn{1}{c|}{\parbox[t]{2mm}{\rotatebox{90}{rider}}} & 
  \multicolumn{1}{c|}{\parbox[t]{2mm}{\rotatebox{90}{car}}} & 
  \multicolumn{1}{c|}{\parbox[t]{2mm}{\rotatebox{90}{trunk}}} & 
  \multicolumn{1}{c|}{\parbox[t]{2mm}{\rotatebox{90}{plants}}} & 
  \multicolumn{1}{c|}{\parbox[t]{2mm}{\rotatebox{90}{traf.}}} & 
  \multicolumn{1}{c|}{\parbox[t]{2mm}{\rotatebox{90}{pole}}} & 
  \multicolumn{1}{c|}{\parbox[t]{2mm}{\rotatebox{90}{garb.}}} & 
  \multicolumn{1}{c|}{\parbox[t]{2mm}{\rotatebox{90}{buil.}}} & 
  \multicolumn{1}{c|}{\parbox[t]{2mm}{\rotatebox{90}{cone.}}} & 
  \multicolumn{1}{c|}{\parbox[t]{2mm}{\rotatebox{90}{fence}}} & 
  \multicolumn{1}{c|}{\parbox[t]{2mm}{\rotatebox{90}{bike}}} & 
  \multicolumn{1}{c|}{\parbox[t]{2mm}{\rotatebox{90}{groun.}}} & 
  \multicolumn{1}{c|}{\parbox[t]{2mm}
  {\rotatebox{90}{mIoU}}}\Tstrut\Bstrut\\
% \cline{3-22}
% & & & & & & & & & & & & & & & & & & & & & \\
\hline
\multirow{2}{*}{SynLiDAR} & 0.5 &
& 55.56 & 52.78 & 35.09 & 23.43 & 71.01 & 22.97	& 31.88 & 30.35 & 67.48 & 21.22 & 24.82 & 10.00	& 78.60 & 40.40\\
% & 0.1
% & - & - & - & - & - & - & - & - & - & - & - & - & - & - \\
% & 0.3
% & - & - & - & - & - & - & - & - & - & - & - & - & - & - \\
% & 0.7
% & - & - & - & - & - & - & - & - & - & - & - & - & - & - \\
& 0.9 &
& 54.59 & 51.67 & 36.06 & 23.55 & 71.40 & 23.56 & 33.41 & 30.48 & 66.8 & 19.25 & 24.71 & 9.74 & 79.05 & 40.33\\
& 0.5 & \cmark
& 55.63 & 51.19 & 37.69 & 21.73 & 72.50 & 18.66 & 36.89	& 23.25 & 68.60 & 21.56 &	26.20 &	9.95 & 79.23 & 40.24\\
\hline
\multirow{2}{*}{AdvSynLiDAR} & 0.5 &
& 55.90	& 34.94	& 11.01	& 14.36	& 27.51	& 14.28	& 18.19 & 10.63 & 52.12 & 5.12 & 26.77 & 16.78 & 71.38 & 27.61\\
% & 0.1
% & - & - & - & - & - & - & - & - & - & - & - & - & - & - \\
% & 0.3
% & - & - & - & - & - & - & - & - & - & - & - & - & - & - \\
% & 0.7
% & - & - & - & - & - & - & - & - & - & - & - & - & - & - \\
& 0.9 &
& 54.80 & 42.37 & 13.56 & 15.91 & 40.45 & 13.24 & 21.50 & 12.89 &	52.68 &	3.96 & 21.63 & 14.57 & 73.12 & 29.28 \\
& 0.5 & \cmark
& 55.70 & 46.36 & 10.99 & 16.51 & 28.18 & 21.43 & 31.52 & 8.54 & 49.27 & 9.77 & 40.00 & 14.80 & 69.89 & 30.30\\
\hline
\hline
\end{tabular}
}
% \smallskip
\label{table:4-2 lidar2poss}
% \vspace{-5mm}
\end{table*}

% \noindent \textbf{Impact of Adversarial Perturbations.}
\subsubsection{Impact of Adversarial Perturbations}
Even minor adversarial perturbations in the source domain lead to significant performance degradation in the target domain across multiple semantic categories. For instance, vegetation IoU in SemanticKITTI drops from 68.72 to 35.90, highlighting the vulnerability of head categories to adversarial attacks. Similarly, categories such as plants in SemanticPOSS also show a considerable drop in IoU. This emphasizes the need for robust cross-domain adaptation techniques as adversarial modifications propagate through the network and deteriorate semantic transferability. 

\subsubsection{Robustness of Long-Tail Classes}
Notably, we observe the impact of adversarial perturbations is less severe on long-tail categories like persons, poles, and bicyclists, which tend to have sparser representations in the dataset. These categories maintain relatively stable segmentation performance compared to head categories, illustrating the inherent distribution imbalance within the source domain and how adversarial perturbations primarily affect the more frequent classes. We observe an improvement in semantic segmentation performance under adversarial attacks by empirically adjusting the selection-perception parameter to 0.9, which increases the focus on long-tail categories. Specifically, the mIoU on the SemanticKITTI dataset improved from 15.28 to 19.87, while on the SemanticPOSS dataset, it increased from 27.61 to 29.28. This adjustment allows long-tail categories to partially recover their segmentation performance while mitigating the performance degradation observed in head categories.

\subsubsection{Effectiveness of AAF}
The Adversarial Adaptation Framework (AAF) significantly mitigates the adverse effects caused by perturbations in the source domain. With AAF, we observe substantial improvements in segmentation performance across both datasets. For example, on the SemanticKITTI dataset, applying the AAF leads to a mIoU improvement from 15.28 (without AAF) to 22.78 (with AAF) under a selection-perc of 0.5. Similarly, for the SemanticPOSS dataset, the mIoU increases from 27.61 to 30.30 with the incorporation of AAF. AAF also demonstrates the ability to stabilize the performance degradation of long-tail categories while improving the overall segmentation. Specifically, in the case of SemanticPOSS, the performance of long-tail categories such as rider and pole is preserved, maintaining performance under adversarial conditions. 
% This is attributed to AAF's ability to better handle domain discrepancies by dynamically adjusting the importance of different semantic categories, especially under adversarial perturbations. 

\begin{figure*}[t]
\begin{center}
   \includegraphics[width=0.85\linewidth]{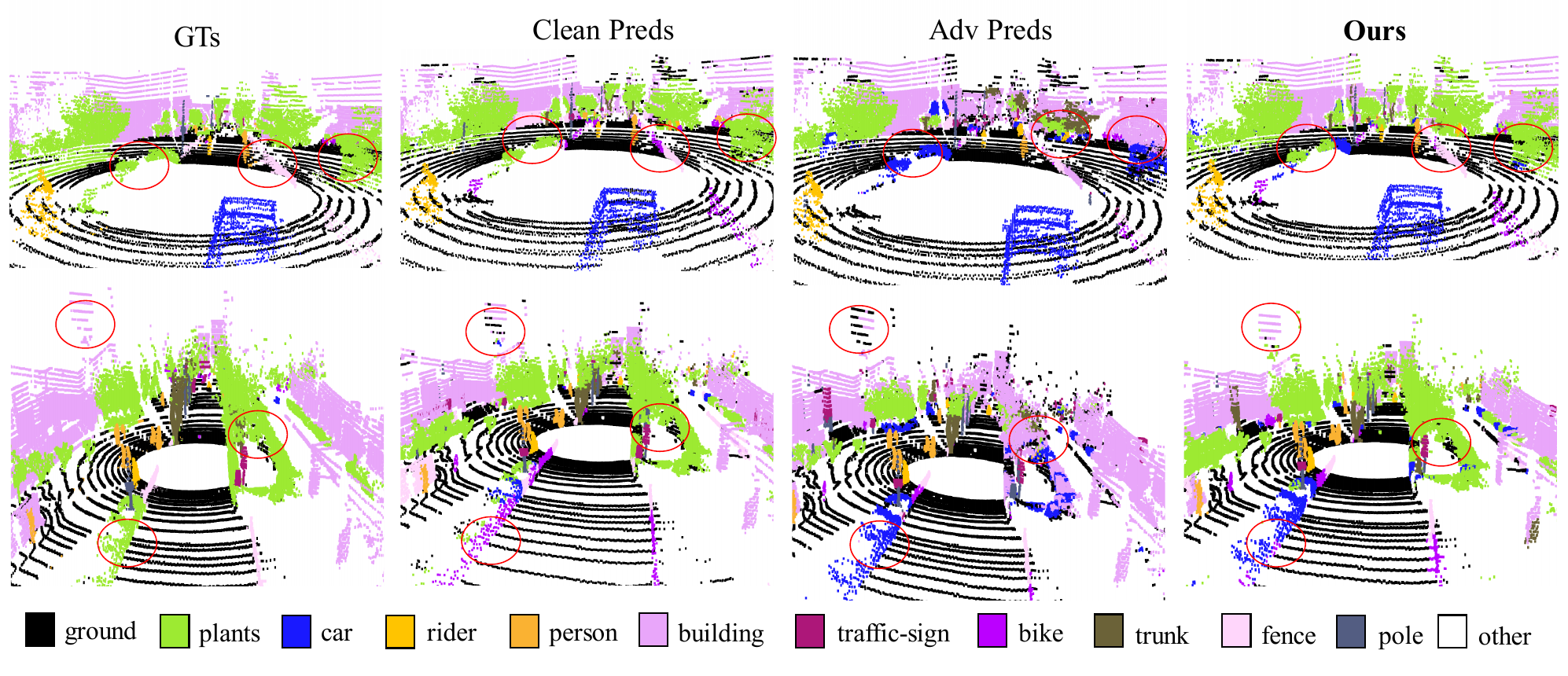}
   % \vspace{-2mm}
\end{center}
    \caption{Visualization of cross-domain adaptation semantic segmentation results on SynLiDAR $\rightarrow$ SemanticPOSS. From left to right: point clouds with the ground truth labels, adaptation with clean SynLiDAR, adaptation with adversarial SynLiDAR, and Ours (adaptation with adversarial SynLiDAR + AAF) }
% \vspace{-5mm}
\label{figure:cross-domain-representations}
\end{figure*}

\subsection{Ablation Studies, Analysis and Discussions}
To further evaluate the efficacy of the AAF framework and justify our design choices, we conduct extensive ablation studies, as shown in Table. \ref{table:4-6-ablation-3}. These studies assess the individual contributions of key components, including the RLT loss during pre-training and the probability decoder branch (Pro.D.) during the cross-domain adaptation phase. We also examine the impact of fine-tuning the model using a small portion of clean source domain data, an essential part of our approach for adapting under adversarial conditions.

% \noindent \textbf{Fine-tuning.} 
\subsubsection{Fine-tuning}
Existing research, such as~\cite{Agarwal_2023_CVPR}, shows that a small amount of clean data could substantially enhance the robustness of the model when faced with adversarial challenges. In our work,  we use 5\% clean source data for training and validation during adaptation with fine-tuning. Only the penultimate convolution and final classification layers are updated, while other parameters remain frozen to retain learned features. Training stops if segmentation performance on the source validation set stagnates for three epochs. As shown in row (d), fine-tuning improves mIoU from 15.28 to 26.17 on SemanticKITTI and 27.75 to 35.66 on SemanticPOSS, highlighting its impact on adversarial robustness.

\subsubsection{RLT Loss}
Incorporating RLT loss without fine-tuning (row (a)) yields substantial improvements, particularly on SemanticKITTI, where the mIoU increases from 15.28 to 22.53.
% This indicates that RLT loss is highly effective in leveraging the underrepresented long-tail categories, which are more robust to adversarial attacks. 
By concentrating on these long-tail categories, RLT loss enables the model to focus more effectively on critical semantic information, enhancing its ability to generalize across domains. Combining RLT loss with fine-tuning (row (e)) further improves the mIoU to 26.24 on SemanticKITTI, highlighting that RLT loss works synergistically with fine-tuning.
% This combination allows the model to adapt effectively to clean data while maintaining adversarial robustness, particularly for long-tail categories.

\begin{table}[t]
% \footnotesize
\center
\caption{Ablation experiments with the components of AAF. RLT Loss is the Robust Long-Tail Loss used in AAF, Pro.D. represents the probability decoder branch with HNPU.}
\resizebox{1.\linewidth}{!}{\noindent
\begin{tabular}{c|c|c|c|c|c}
\hline
\hline
  \multirow{2}{*}{Methods} & 
  \multirow{2}{*}{Fine-tuning} &
  \multirow{2}{*}{RLT Loss} &
  \multirow{2}{*}{Pro.D.} &
  \multicolumn{2}{c}{mIoU} \Tstrut\Bstrut\\
  \cline{5-6}
  & & & & SemanticKITTI & SemanticPOSS \Tstrut\Bstrut\\
\hline
CosMix
& - & - & - & 15.28 & 27.75\Tstrut\\
\hline
(a)
&  & \cmark & & 22.53 & 28.11\Tstrut\\
(b)
&  &  & \cmark & 20.07 & 30.14\Tstrut\\
(c)
&  & \cmark & \cmark & 22.78 & 30.30\Tstrut\\
(d)
& \cmark &  &  & 26.17 & 35.66\Tstrut\\
(e)
& \cmark & \cmark &  & 26.24 & 35.81\Tstrut\\
(f)
& \cmark &  & \cmark & 26.43 & 36.45\Tstrut\\
(g)
& \cmark & \cmark & \cmark & \textbf{26.89} & \textbf{37.60}\Tstrut\\
\hline
\hline
\end{tabular}}
% \smallskip
% \vspace{-3mm}
\label{table:4-6-ablation-3}
\end{table}

\subsubsection{Probability Decoder Branch (Pro.D.)}
Introducing the Probability decoder branch (row (b)) results in clear performance gains, particularly on SemanticPOSS, where the mIoU rises from 27.75 to 30.14. When used in conjunction with RLT loss (row (c)), Pro.D. demonstrates its complementary nature, further enhancing the mIoU to 30.30 on SemanticPOSS. Additionally, when all components—fine-tuning, RLT loss, and Pro.D.—are combined (row (g)), the model achieves the best overall performance, with mIoU of 26.89 on SemanticKITTI and 37.60 on SemanticPOSS. This result underscores the effectiveness of integrating multiple strategies to maximize model robustness under adversarial conditions.

Fig.~\ref{figure:cross-domain-representations} visualizes the cross-domain adaptation results under adversarial source domain, highlighting how the AAF framework restores the semantic structure. Recovery in head classes (e.g., `building', `plants') benefits from fine-tuning on clean data, while region-level consistency is mainly attributed to the HNPU strategy. Notably, in some red-circled areas, AAF even surpasses clean predictions in semantic completeness.

\section{Conclusions}
In this paper, we explore the robustness of UDA frameworks under source domain attacks. To our knowledge, this paper is the first paper on such exploration. To benchmark and investigate the challenges of 3D point cloud UDA under point cloud attacks, we propose a novel point cloud adversarial sample generation method to contaminate the dataset, resulting in the new AdvSynLiDAR dataset. Further, we introduce the AAF framework to address the performance degradation by introducing the Robust Long-Tail loss (RLT loss) and designing a decoder-branch-based approach. Experiments show that our proposed AAF framework enhances the model's robustness and ensures effective semantic segmentation in cross-domain tasks.

{\small
\bibliographystyle{IEEEtran}
\bibliography{ieee}
}

% \newpage

 

% \end{IEEEbiography}

\vfill

\end{document}